# Semantic Oriented Agent based Approach towards Engineering Data Management, Web Information Retrieval and User System Communication Problems

Zeeshan Ahmed and Detlef Gerhard

Mechanical Engineering Informatics and Virtual Product Development Division (MIVP), Vienna University of Technology Austria

Telephone: + 43 1 58801 30726 & Fax: + 43 1 58801 30798

Email: {zeeshan.ahmed, detlef.gerhard}@tuwien.ac.at & Web: www.mivp.tuwien.ac.at

Address: Getreidemarkt 9/307 1060 Wien Austria

**Abstract**

The four intensive problems to the software rose by the software industry .i.e., User System Communication / Human Machine Interface, Meta Data extraction, Information processing & management and Data representation are discussed in this research paper. To contribute in the field we have proposed and described an intelligent semantic oriented agent based search engine including the concepts of intelligent graphical user interface, natural language based information processing, data management and data reconstruction for the final user end information representation.

## 1. Introduction

These days software industry is facing four major problems related to the software product development and data management .i.e., User System Communication / Human Machine Interface, Meta Data extraction, Information processing & management and Data representation. Within the User system communication user can be natural or a digital system. In case of a digital system, the input data will be the electronic or digital and most probably be defined in limits which can easily be recognized, analyzed and processed by the system. Whereas if the system is natural then the input data will most probably be logical and natural language based which is quite difficult to handle because of the high complexity level depending on the random and variable change. To process any natural language based instruction its very important to first understand the semantic of instruction, and to understand the semantic Meta data has to be extracted from data, hidden information must be analyzed, processed and managed. In the end, the final output must be presented in bold, clear and understandable format.

Unfortunately regardless of having enough research and development in graphical user interface development but still right now there is no as such intelligent graphical user interface developed which claims of having the ability to intelligently communicate with user and flexible enough so then it can easily be handled and redesigned according to the need by user itself. Moreover if we go in to the details of Meta Data Extraction and Information processing & management problems, then we come to know that there is nothing like a search engine available which can completely extract the Meta data from data, analyze the semantic of any natural language based instructions using the given context of a domain specific problem, process, manage and present resultant data to natural user in bold, clear and understandable format.

In this paper focusing on above discussed problems in section 2 we present an architecture based conceptual solution. Moreover briefly describe its components in sections 2.1, 2.2, 2.3 and 2.4 of the paper. In the end we conclude the discussion in section 3 and present some future aspects in section 4 of the paper.

## 2. ISOAS

Keeping eye on above already discussed four major and very important aspects to be considered before and during the development of most of the software products .i.e., User System Communication / Human Machine Interface, Meta Data Extraction, Information Processing Management and Data Presentation we have proposed a new approach Intelligent Semantic Oriented Agent Based Search (I-SOAS) as the extended version of Semantic Oriented Agent Based Search (SOAS) [1, 2, 3]. The proposed architecture of I-SOAS is consists of four main sequential iterative components .i.e., Intelligent User Interface (IUI), Information Processing (IP), Data Management (DM) and Data Representation (DR) as shown in Figure 1.

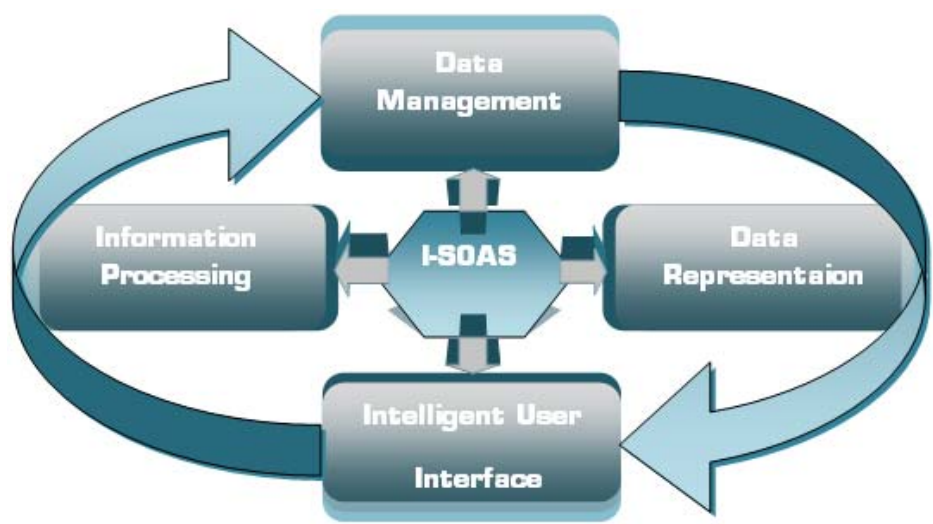

**Figure 1.** Intelligent Semantic Oriented Agent Based Search (I-SOAS)

Every product whether is software or hardware has a user interface for user system communication, here in our proposed architecture of I-SOAS the first component is also the User Interface .i.e., IUI. We have tried to improve concept of User Interface to Intelligent User Interface (IUI) described in detail in section 2.1 of the paper. Considering the concepts of compiler construction of any software programming language we have tried to introduce a new concept of information processing and semantic modelling discussed in detail in section 2.2 of paper. Then following the already available well matured data warehousing concepts to store and manage data we have designed our third component for data management, briefly described in section 2.3 of paper. Then to reconstruct the final output of the system in user's understandable format, we have designed the last component of I-SOAS presented in section 2.4 of paper.

## 2.1 Intelligent User Interface (IUI)

Intelligent User Interface (IUI) is the first component of I-SOAS with respect to the user and system communication's point of view. This component is responsible for the intelligent communication between user and the system. IUI is proposed as an intelligent dynamic user interface which is be capable of first analyzing the source of input, forwarding inputted data for further processing and responding back to user with end results in the same input format. Moreover IUI is supposed to be flexible enough so then it can be redesigned by user itself.

To implement the IUI system architecture as shown in Figure.2 IUI is divided into two main categories .i.e., Graphical User Interface and Communication Sources. In Communication Sources first the corresponding user is identified to enable the correct communication mode. If it is a digital system then electronic data communication mode is enabled and if it is natural system then natural language based communication mode is enabled. Whereas the Graphical User Interface is consists of the concept of three sub categories .i.e., Intelligent, Flexible and Agent to intelligently handle the user's requests, provide multiple options to redesign the graphical user interface according to the ease of the user by user itself and perform internal architectural component's agent based communication.

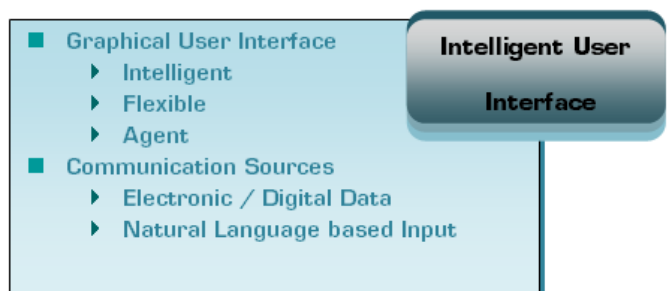

**Figure 2.** I-SOAS – Intelligent User Interface (IUI)

## 2.2 Information Processing (IP)

Information Processing unit is the most important component of the I-SOAS, the quality of its performance depends upon the accuracy in the results of this component. The overall job of IP is divided into five main iterative sequential steps .i.e., data reading, tokenization, parsing, semantic modelling and semantic based query generation as shown in Figure.3. Each step requires intensive and intelligent effort in development. The main concept behind the organisation of these five steps is to first understand the semantic hidden in the context of natural or digital set of instructions and generate a semantic information process able model for the system's own understanding and information processing.

In the first step Data Reader, retrieved data from IUI is just read and organised without performing any analytical intelligent operation except initial prioritization of instructions in to a list. Then in the second phase Data Tokenizer (Lexer), instructions are tokenized in to the possible number of tokens which are then treated in the third step Data Parser for parsing and semantic evaluation with respect to the grammar of used natural or digital language. If someone has the idea of compiler construction then he can better understand the process of tokenization and parsing because this works on almost the same principles which are normally followed by any other compiler of any programming language.

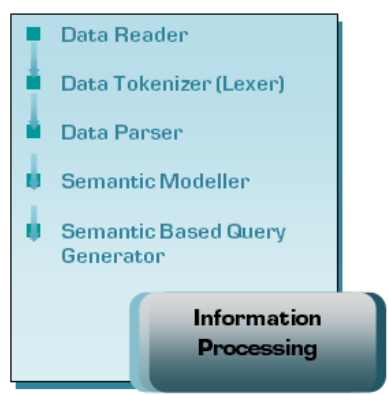

**Figure 3.** I-SOAS - Information Processing (IP)

Then in the fourth and the most important phase of IP Semantic Modeller, the tokenized and semantically evaluated information is used and reorganised in Meta data based semantic model. The main purpose of the creation of Semantic Modeller is first to filter the irrelevant semantic less helping grammar based information from the actual information based context and then create a semantic model which our own system can easily analyze and process. Then in the last phase Semantic Based Query Generator based on the output of Semantic Modeller generates a new query used further for the extraction of desired result in next component Data Management.

## 2.3 Data Management (DM)

As the concept of data management is concerned, data management is based on two main principles .i.e., storing data and managing data. In I-SOAS Data Management is playing the heaviest component's role due to the size, because the size of this component will increase with the increase of data with the passage of time. This component is responsible for two main functions .i.e.; Semantic based Query Processing and Data Management.

Semantic based query build in IP is treated by Semantic based Query Processor to generate

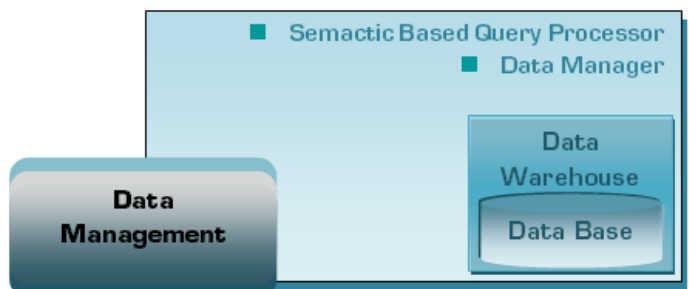

**Figure 4.** I-SOAS - Data Management (DM)

SQL query to run in to data warehouse to extract the required relevant information. The job of Data Manager is to manage the process of SQL query building, data extraction and creation of new indexes and storage based on newly retrieved information.

## 2.4 Data Representation (DR)

Data Representation is the last component of I-SOAS carrying one of the important jobs of responding back to the user with finalized results. This component consists of six sub components .i.e., Information Retriever, Information Reader, Information Tokenizer, Information Parser, Information Reconstructor and Presenter. The job of this component somehow is relevant to the job of IP, but major difference is of handling data and information. IP treats data to process but DR treats information.

Required extracted and managed information from Data Manager is passed to DR using Information Retriever, which simply read and organised by Information Reader without performing any analytical intelligent action except prioritization of informative statements. Then using Information Tokenizer and Information Parser statements are tokenized and parsed and using Information Reconstructor. Finalized formatted information is build in user's used natural language. Finally Presenter presents the resultant information to IUI to respond back to the user.

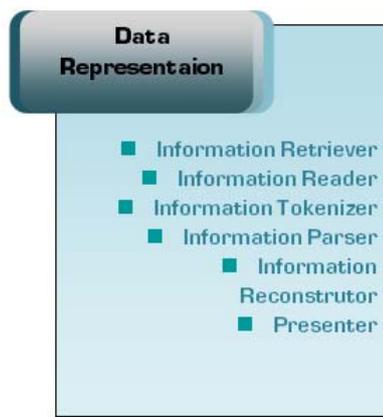

**Figure 5.** I-SOAS - Data Representation

## 3. Conclusion

In this paper we addressed four major software product development and data management related problems .i.e., User System Communication / Human Machine Interface, Meta Data extraction, Information processing & management and Data representation. Then describing our own contribution towards the field we have presented an approach and its conceptual design.

## 4. Future Work

As this is an ongoing in process research, in future, we are aiming to focus on the in depth architectural details, mathematical or algorithmic experiments and currently available tools and technologies for the development of I-SOAS as real time software application. We are confident to present detailed, concrete and optimized results using real data sets during and after the implementation of I-SOAS. Moreover, as we are promoting I-SOAS as a dynamic concept, so we will also try to prove the importance of I-SOAS in different fields.

## References

1. A. Zeeshan and D. Gerhard, "An Agent based Approach towards Metadata Extraction, Modelling and Information Retrieval over the Web", *In the proceedings of First International Workshop on Cultural Heritage on the Semantic Web in conjunction with the 6th International Semantic Web Conference and the 2nd Asian Semantic Web Conference 2007*, Busan Korea, 12-15 November 2007

2. A. Zeeshan and D. Gerhard, "Designing a Dynamic Components and Agent based Approach for Semantic Information Retrieval", *In the proceedings of 6th CIIT Workshop on Research in Computing (CWRC Fall 2007) COMSATS Institute of Information Technology*, Abbottabad Pakistan, 27 October 2007

3. A. Zeeshan and D. Gerhard, "Personal Assistant towards Semantic Information Retrieval", *In the proceedings of Fifth International Workshop on Ontologies and Semantic Web for E-Learning (SWEL'07),13th International Conference on Artificial Intelligence in Education (AIED 2007)*, Los Angeles USA, 9-13 July 2007